\documentclass[runningheads]{llncs}
\usepackage{graphicx}

\usepackage{dsfont}
\usepackage{bbm}
\usepackage{booktabs}
\usepackage{multirow}
\usepackage{siunitx}

\usepackage{amsmath,amssymb} 
\usepackage{color}
\usepackage[width=122mm,left=12mm,paperwidth=146mm,height=193mm,top=16mm,paperheight=237mm]{geometry}
\begin{document}
\pagestyle{plain}
\mainmatter

\title{SiamVGG: Visual Tracking using Deeper Siamese Networks} 


\author{Yuhong Li, Xiaofan Zhang, Deming Chen \\
        \textit{\{leeyh,xiaofan3,dchen\}@illinois.edu}
}
\institute{University of Illinois at Urbana-Champaign}

\maketitle

\begin{abstract}
Recently, we have seen a rapid development of Deep Neural Network (DNN) based visual tracking solutions. Some trackers combine the DNN-based solutions with Discriminative Correlation Filters (DCF) to extract semantic features and successfully deliver the state-of-the-art tracking accuracy. However, these solutions are highly compute-intensive, which require long processing time, resulting unsecured real-time performance. To deliver both high accuracy and reliable real-time performance, we propose a novel tracker called SiamVGG\footnote{https://github.com/leeyeehoo/SiamVGG}. It combines a Convolutional Neural Network (CNN) backbone and a cross-correlation operator, and takes advantage of the features from exemplary images for more accurate object tracking.
The architecture of SiamVGG is customized from VGG-16 with the parameters shared by both exemplary images and desired input video frames. 
%
We demonstrate the proposed SiamVGG on OTB-2013/50/100 and VOT 2015/2016/2017 datasets with the state-of-the-art accuracy while maintaining a decent real-time performance of 50 FPS running on a GTX 1080Ti. Our design can achieve 2\% higher Expected Average Overlap (EAO) compared to the ECO~\cite{danelljan2017eco} and  C-COT~\cite{danelljan2015convolutional} in VOT2017 Challenge.

\end{abstract}

\section{Introduction}
\label{sec:Introduction}
Visual object tracking is one of the most fundamental topics in computer vision. Building trackers with both high accuracy and reliable real-time performance is always a challenging problem. In the tracking problems, a bounding box is given of an arbitrary object in the first frame, and the goal is to report the location of the same target in following frames. With the extremely high practicability, there are growing number of tracking-related applications, which can be easily found in surveillance systems, Unmanned Aerial vehicles (UAVs), and self-driving cars, etc. However, higher demands of accuracy and real-time performance are proposed by real-life applications since trackers can be easily distracted by the movement of targeted and surrounding objects in real scenarios, such as motion changes, illumination changes, and occlusion issues, and real-time capability can guarantee the satisfaction of input video frame rate. 
However, it is hard to satisfy real-time tracking while using DNN-based trackers with high computation complexity and long processing latency. Although we can take advantage of hardware accelerators using GPUs~\cite{nvdawhitepaper1,zhao2012real,nvdawhitepaper2,li2020edd}, FPGAs~\cite{zhang2017high,zhangdnnbuilder,xu2020autodnnchip,chen2016platform,liu2011real,zhuge2018face,zhang2020skynet,he2009novel}, or ASICs~\cite{isscc_2016_chen_eyeriss,han2016eie,jouppi2017datacenter}, the tradeoffs between accuracy and real-time performance of DNN-based trackers using customized accelerators are not yet fully investigated.

Recently, a Discriminative Correlation Filters (DCF) based approach has demonstrated its promising real-time performance in tracking applications. This approach transforms convolutional operations from time-domain to frequency-domain so that it largely improves the computation efficiency and speed. One typical example is a tracker called MOSSE~\cite{bolme2010visual}, which can process up to 700 frames per second but it fails to provide acceptable accuracy and requires additional hand-crafted features to recognize the targeted object.

Another solution relies on DNNs to capture features of the targeted object. DNNs have demonstrated their potentials of feature extractions especially in object classification~\cite{krizhevsky2012imagenet}, detection~\cite{redmon2017yolo9000}, and saliency prediction~\cite{li2018csrnet} tasks. By using DNNs, more high-dimensional features can be captured to significantly improve the object tracking algorithms. With the integration of convolutional neural networks (CNNs) as the feature extractors, DCF based trackers begin to show improving performance on various tracking datasets regarding accuracy. However, most of these trackers still cannot meet the real-time requirement (e.g., $\geq$30 FPS)~\cite{danelljan2017eco}, because every incoming frame needs to go through all layers of the CNN to get the up-to-date features, resulting in a time-consuming procedure. Another problem of DNN-based object trackers is that the feature extractors (the CNN part) are not originally designed and trained for tracking tasks. For example, in the ECO tracker~\cite{danelljan2017eco}, the feature extractor is the first (Conv-1) and last (Conv-5) convolutional layer of the VGG-m, which are pretrained on image classification datasets but not on tracking datasets. Without the dedicated end-to-end training for tracking tasks, the tracking-oriented DNN models may be affected by the training noises and perform poorly. Some models with online learning ability may overcome this problem by fine-tuning the network parameters during tracking, such as using stochastic gradient descent (SGD), but the online learning approach does not satisfy the real-time requirement either.

\begin{figure}[t]
\centering
\includegraphics[width=3cm, height=2.6cm]{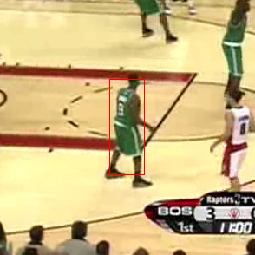}
\includegraphics[width=3cm, height=2.6cm]{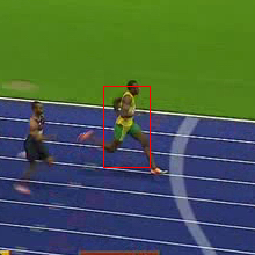}
\includegraphics[width=3cm, height=2.6cm]{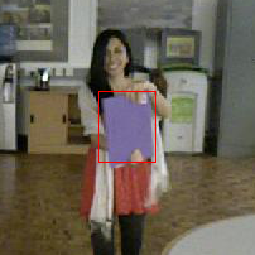}\\
\includegraphics[width=3cm, height=2.6cm]{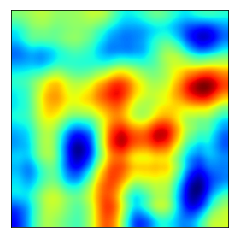}
\includegraphics[width=3cm, height=2.6cm]{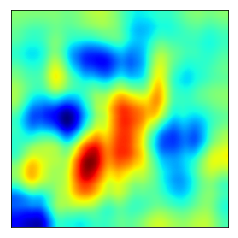}
\includegraphics[width=3cm, height=2.6cm]{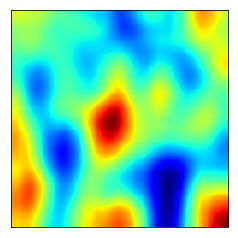}\\
\includegraphics[width=3cm, height=2.6cm]{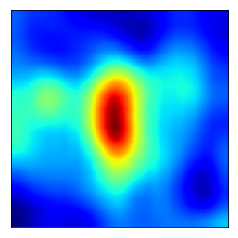}
\includegraphics[width=3cm, height=2.6cm]{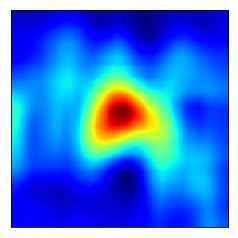}
\includegraphics[width=3cm, height=2.6cm]{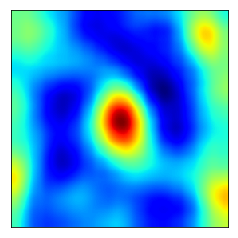}

\caption{The first row presents the targeted objects (with bounding boxes) in VOT2017, while the second and the third rows represent the  
score maps of the possible locations of the targeted object using SiamFC and the proposed SiamVGG, respectively.
}
\vspace{-18pt}
\label{fig:comparison}
\end{figure}

To balance accuracy and speed, offline DNN-based trackers have potential to deliver promising solutions. Among them, the approaches which consider tracking problems as similarity learning problems get the highest attentions recently. These solutions use Siamese Network, such as the fully-convolutional Siamese Network (SiamFC)~\cite{bertinetto2016fully}, to learn the similar parts of the targeted objects out of the input frames. Since this solution is end-to-end trainable, it is easier to train on object detection datasets.
As a result, tracking solutions with Siamese Networks can deliver high accuracy without any fine-tuning or online updating process. However, the discrimination ability of these solutions are highly relying on the feature extraction capabilities of the Siamese Network. For example, our proposed design, SiamVGG (using modified VGG-16), performs much better than the SiamFC~\cite{bertinetto2016fully} (using AlexNet), where more details are shown in Figure~\ref{fig:comparison}. 
Larger portion of the red color area in the second row represents that SiamFC can be easily disturbed by the similar objects or even backgrounds, resulting in uncertain output predictions compared to the third row with a better Siamese Network.

In this paper, we propose a new approach for object tracking, named SiamVGG, to reduce the major drawback (weak discrimination capability) of the current Siamese Network based methods. We adapt more advanced networks for better discrimination capability and eventually improve the proposed Siamese Network based tracker. In our experiments, we notice that not all the networks are suitable for the Siamese structure and we choose the VGG-16~\cite{simonyan2014very}, which shows the best performance, as the backbone CNN, and it is trained on both ILSVRC dataset~\cite{krizhevsky2012imagenet}, and Youtube-BB dataset~\cite{real2017youtube}. We evaluate the proposed SiamVGG in VOT 2015, VOT 2016, VOT 2017, OTB-100, OTB-50, and OTB-2013 datasets~\cite{WuLimYang13,Kristan2017a,VOT_TPAMI}. The proposed SiamVGG delivers the state-of-the-art performance without tiring fine-tuning efforts for hyperparameters. In addition, the proposed network is very compact so that it can reach 50 FPS for most of the real-time applications.

The rest of the paper is organized as follows. Related works are introduced in Section~\ref{sec:Related}.
The proposed architecture and corresponding configurations are introduced in Section~\ref{sec:proposed}. Section~\ref{sec:Experiments} presents the experimental results on six datasets and Section~\ref{sec:conclusion} concludes this paper.

\section{Related Work}
\label{sec:Related}

\subsection{Using CNNs as Feature Extractors}

The features captured by CNNs can enhance the performance of traditional tracking algorithms by bringing sufficient high-level semantic representations to locate the desired targets. One of the successful examples is called DeepSRDCF~\cite{danelljan2015convolutional}, which takes advantage of the features extracted by CNN combined with correlation filters. Similar methods, such as C-COT~\cite{danelljan2016beyond} and ECO~\cite{danelljan2017eco}, employ continuous convolution operators to enhance the feature extraction of their trackers. Although these methods can reach the state-of-the-art tracking accuracy, they are unable to deliver high enough frame rate (Frame per second) to support real-time object tracking, especially when input images become large.

\subsection{Siamese Network Based Tracker}
To overcome the low-frame-rate challenge, recently published papers, such as SiamFC~\cite{bertinetto2016fully}, start to solve the tackling problem through similarity learning. They exhaustively search all possible locations and
choose the candidate with the maximum similarity to the past appearance of the object. The goal of SiamFC is to find the potential area of the targeted objects in the following frames by comparing with the exemplary images. In this approach, the tracker does not need to perform the online parameter update, which helps to satisfy the frames rate requirement of real-time detection. This work proposes a network that shares the weights for both search and exemplary phases.
There are also a growing number of SiamFC-like structures proposed recently. Among them,  EAST~\cite{huang2017learning} tries to speed up the tracker by learning an agent to decide whether to locate objects with high confidence on an early layer. DSiam~\cite{guo2017learning} attempts to adjust the exemplars by adding online updating, while SA-Siam~\cite{he2018twofold} implements a two-branch Siamese Network with one branch for semantic and the other for appearance. In addition, RASNet~\cite{wang2018learning} introduces three different attention mechanisms to enhance the object segmentation performance, and SiamRPN~\cite{li2018high} integrates the region proposal network as the backend to improve the scaling speed. However, most of these methods still use AlexNet as their backbone, which is not good enough for extracting the underlying semantic features for the targeted objects.

\section{Proposed Solution}
\label{sec:proposed}

The main idea of our design is to deploy a stronger DNN to capture more detailed semantic features and to combine it with the advantages of using the Siamese Network. In this section, we first introduce the proposed network architecture, and then we present the corresponding training methods. Finally, we illustrate the tracking configurations in our design. 

\subsection{SiamVGG Architecture}

Following the idea of the Siamese Network, we choose the modified VGG-16 as the backbone because of its strong transfer learning ability shown in other tasks (e.g., segmentation~\cite{long2015fully}, crowd counting~\cite{li2018csrnet}, etc.) 
and its straightforward structure for easy implementation. We also follow the base design strategy from SiamFC and use it as our baseline design for further comparisons.
 
\subsubsection{Fully-convolutional Siamese Network}

We first introduce the Siamese Network using fully-convolutional operations. Assuming $L_{\tau}$ is the translation operator, we have $(L_{\tau}x)[u] = x[u - {\tau}]$ with $x\in \mathbb{R}^{W \times H}$. The symbol $u$ represents the position $(i,j) \in \mathbb{Z}^2$, where $\forall i \in [1,W]$ and $\forall j \in [1,H]$. 
The symbol $\tau$ ($\tau \in \mathbb{Z}$) is an offset and we have $u-\tau = (i-\tau,j)$ or $u-\tau = (i,j-\tau)$. So, $i-\tau \in [1,W]$ or $j-\tau \in [1,H]$. When the mapping function $h$ for input signals is a fully-convolutional function with integer stride $k$, it needs to meet the following Equation (1), for translation $\tau$ in the valid region of both inputs and outputs.

\begin{align}
h(L_{k\tau}x) = L_{\tau}h(x)
\end{align}

By using the fully convolutional Siamese Network, we need to feed in two input images and compute their cross-correlation, $f(z, x)$, which is defined as Equation (2), where $z$ and $x$ represent the exemplary image and the search image, respectively. In Equation (2), $\varphi$ represents a convolutional embedding function (e.g., CNNs) and $b\mathbbm{1}$ denotes the bias with value $b \in \mathbb{R}$. In our experiments, we notice that the $b\mathbbm{1}$ contributes nearly nothing to the final performance, so that we remove it in our design for a more compact structure.

\begin{align}
f(z, x) = \varphi(z) \ast \varphi(x) + b\mathbbm{1}
\end{align}

\begin{figure}
\centering
\includegraphics[height=4cm]{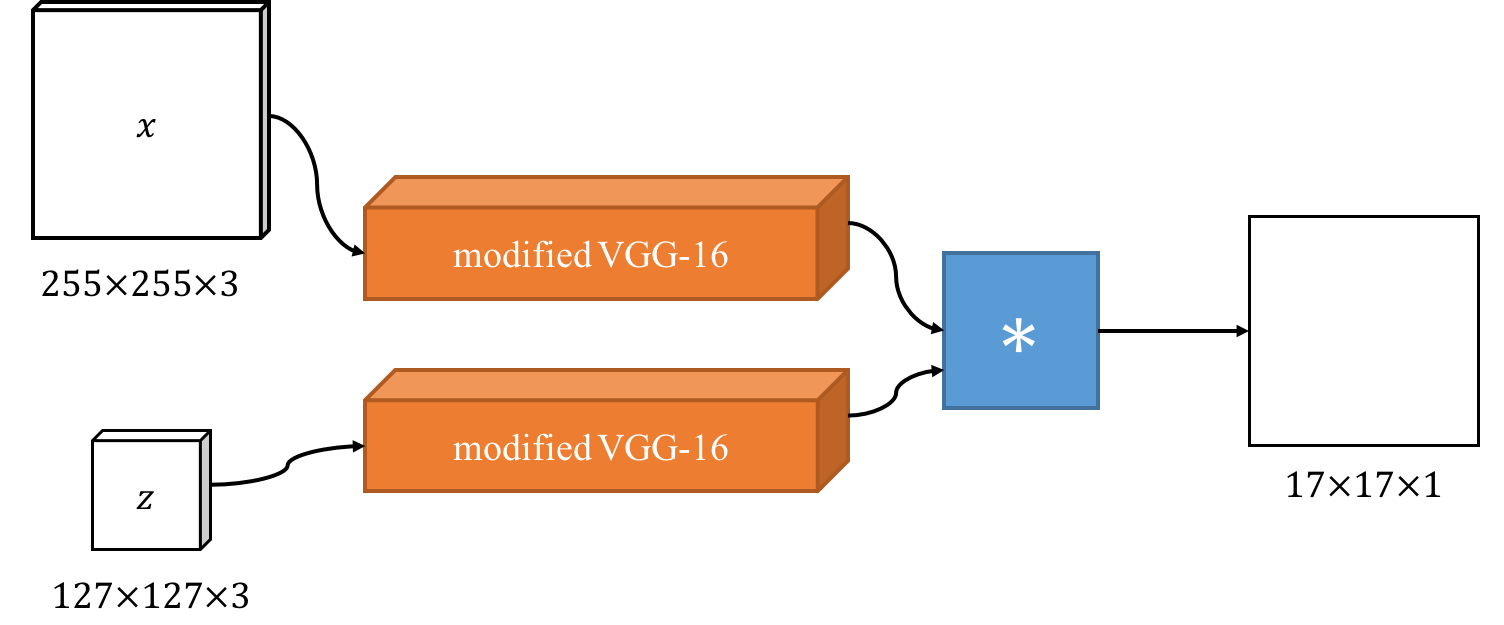}
\vspace{-8pt}
\caption{Architecture of the proposed SiamVGG with VGG-16 based Siamese Network. }
\label{fig:model}
\end{figure}

Figure~\ref{fig:model} shows the whole network structure of our proposed design with two inputs ($x$ for the search image and $z$ for the exemplary image) and one output for the score map after running cross-correlation between two inputs. The generated output map indicates the similarity information between the current search image and exemplary image. 

During tracking, the search image is centered at the previous position of the target which the network predicts. By using the bicubic interpolation, the score map is re-scaled to a specific size which equals to the original score map size multiplied by the down-sample rate of the network. The position of the targeted object is obtained according to generated score map (where the maximum score means the best possible position). Multiple scales are searched in a single inference by stacking a mini-batch of scaled images.

One interesting fact is that the commonly used padding operation in CNNs fails to show any benefits in the fully-convolutional Siamese Network. One major reason comes from the noise introduced by padding operations, which deteriorates the quality of the output feature maps after pooling and leads to imperfect score maps after running cross-correlation. Eventually, the trackers using Siamese Network with padding suffer accuracy loss and the most intuitive way is to remove padding. However, the feature maps from the following layers would shrink as no padding operations are applied, which may limit the depth of network to prevent the feature map from becoming too small and eventually damage the network quality. 
As a result, the network depth needs to be carefully configured to avoid the above problem. 

\subsubsection{Network Configuration}

We illustrate the proposed network configuration in Table~\ref{table:siamvgg}. By removing the fully-connected layers, we keep the first 10 layers from VGG-16 and 
make the last layer a $1 \times 1$ kernel as the output. There are two reasons for us to choose VGG-16 as the backbone network. The first reason is that we need to implement a CNN without using padding, which prevents us from using most of the up-to-date CNNs, such as ResNet (with shortcut connections between layers) and GoogLeNet (with Inception modules), since padding operations are required in these networks. 
The second reason is due to the great adaptability of VGG-like networks which allows us to adapt them from image classification to object tracking. To be more specific, these networks can be pre-trained on classification datasets, and then adapted to tracking tasks. 
\setlength{\tabcolsep}{4pt}
\begin{table}[t]
\begin{center}
\caption{The backbone architecture of SiamVGG. All the convolutional layers are integrated with ReLU except the last one working for generating outputs. `MP' stands for the maxpooling layer. The channel map indicates the number of output and input channels using the format $output channel \times input channel$.}
\label{table:siamvgg}
  \begin{tabular}{ccccccc}
    \toprule
    
       \quad& \quad & \quad
      & \quad &
      \multicolumn{3}{c}{Activation Size} \\
     Layer & Kernel Size & Chan. Map & Stride & For Exemplar & For Search & Chan. \\
      \midrule
    \quad  & \quad        & \quad            & \quad & $127 \times 127$ & $255 \times 255$ & $\times 3  $ \\
    CONV1  & $3 \times 3$ & $64 \times 3$    & 1     & $125 \times 125$ & $253 \times 253$ & $\times 64 $ \\
    CONV2  & $3 \times 3$ & $64 \times 64$   & 1     & $123 \times 123$ & $251 \times 251$ & $\times 64 $ \\
    MP1    & $2 \times 2$ & \quad            & 2     & $61  \times 61 $ & $125 \times 125$ & $\times 64 $ \\
    CONV3  & $3 \times 3$ & $128 \times 64$  & 1     & $59  \times 59 $ & $123 \times 123$ & $\times 128$ \\
    CONV4  & $3 \times 3$ & $128 \times 128$ & 1     & $57  \times 57 $ & $121 \times 121$ & $\times 128$ \\
    MP2    & $2 \times 2$ & \quad            & 2     & $28  \times 28 $ & $60  \times  60$ & $\times 128$ \\
    CONV5  & $3 \times 3$ & $256 \times 128$ & 1     & $26  \times 26 $ & $58  \times  58$ & $\times 256$ \\
    CONV6  & $3 \times 3$ & $256 \times 256$ & 1     & $24  \times 24 $ & $56  \times  56$ & $\times 256$ \\
    CONV7  & $3 \times 3$ & $256 \times 256$ & 1     & $22  \times 22 $ & $54  \times  54$ & $\times 256$ \\
    MP3    & $2 \times 2$ & \quad            & 2     & $11  \times 11 $ & $27  \times  27$ & $\times 512$ \\
    CONV8  & $3 \times 3$ & $512 \times 256$ & 1     & $9   \times 9  $ & $25  \times  25$ & $\times 512$ \\
    CONV9  & $3 \times 3$ & $512 \times 512$ & 1     & $7   \times 7  $ & $23  \times  23$ & $\times 512$ \\
    CONV10 & $3 \times 3$ & $512 \times 512$ & 1     & $5   \times 5  $ & $21  \times  21$ & $\times 512$ \\
    CONV11 & $3 \times 3$ & $256 \times 512$ & 1     & $5   \times 5  $ & $21  \times  21$ & $\times 256$ \\
    \bottomrule
  \end{tabular}
\end{center}
\end{table}
\setlength{\tabcolsep}{1.4pt}

\subsection{Training Method}

In this section, we provide details of our training method. By using regular CNN components (including only convolutional layers, Relu, and maxpooling layers), our proposed SiamVGG is easy to implement and fast to deploy.

\subsubsection{Generating Ground Truth}

To train the SiamVGG, we use ILSVRC and Youtube-88 datasets, which cover around 4,000 videos annotated frame-by-frame and more than 100,000 videos annotated every 30 frames, respectively. Because ILSVRC contains more fine-grained information while Youtube-BB covers the coarse-grained information, we set the learning ratio between these two datasets as $1 : 5$. 

We randomly pick two frames from the same video sequence and assemble them into a pair of the search image and the exemplary image. In our design, the size of search image is $255 \times 255$ while the size of the exemplary image is $127 \times 127$. To keep all images as squares, we add context margin on top of the original images following the Equation (3) and (4). Assuming the original image size $(w, h)$, and the context margin $p$, we can calculate the scale factor $s$ in Equation (3). 
Regarding the exemplary images, $A = 127 \times 127$ and the context margin is $p = (w + h)/4$. So, in the original frame, the target object is centered at the original center of the bounding box with side length $L$ calculated using Equation (4). The search images are $255 \times 255$, which means that they are centered at the original bounding box center with $2L$ side length. 
\begin{align}
    s(w + 2p) \times s(h + 2p) = A
\end{align}
\begin{align}
    L = \sqrt{(w + 2p) \times (h + 2p)}
\end{align}

\begin{figure}
\centering
\includegraphics[width=.3\textwidth]{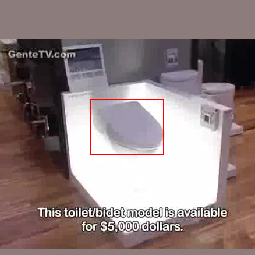}\hfill
\includegraphics[width=.3\textwidth]{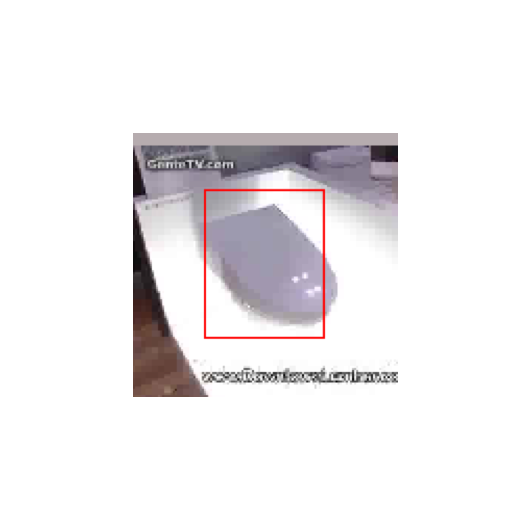}\hfill
\includegraphics[width=.3\textwidth]{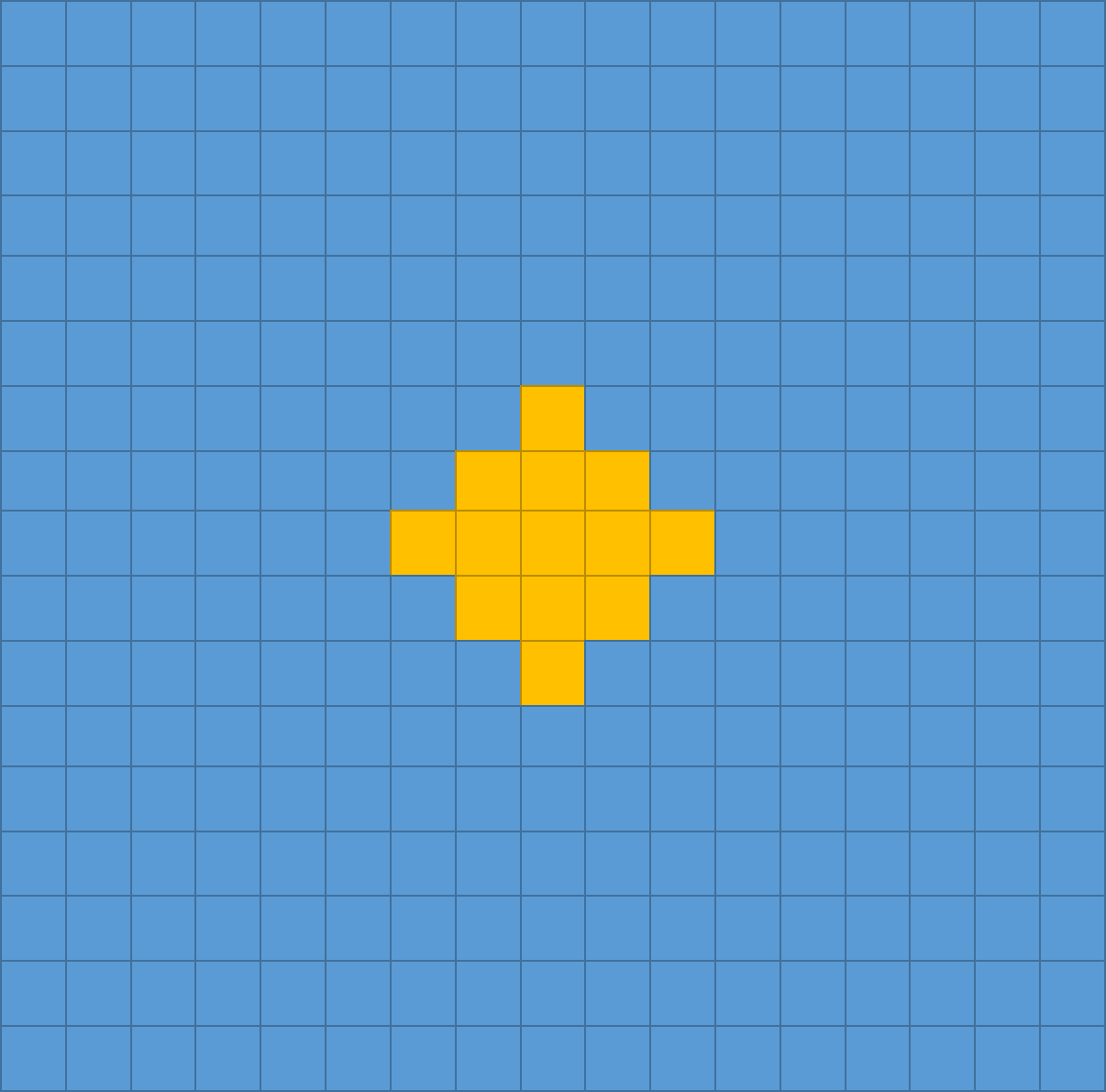}
\caption{The search image (left) with size  $255 \times 255$, the exemplary image (middle) with sized $127 \times 127$, and the ground truth score map (right) with size $17 \times 17$.}
\label{fig:xzgt_example}
\end{figure}

Since the output of the proposed structure in Figure~\ref{fig:model} is $17 \times 17$, we set the ground truth of the score map to $17 \times 17$. By using Equation (5), we generate the elements in the score map, which are considered as positive examples if the targeted objects are located in the center within specific radius R using Manhattan distance. $y[u]$ represents the elements of the ground truth while $c$ is the center of the score map. In the example in Figure~\ref{fig:xzgt_example}, we assume $R = 2$ and coefficient $k = 1$.

\begin{align}\label{eq:y}
y[u] = \left\{\begin{matrix}
+1 & \text{if }k\left \| u - c \right \|\leq R\\ 
-1 & \text{otherwise }.
\end{matrix}\right.
\end{align}

For data augmentation, we randomly stretch the search image on a small scale, from $1.04^{-3}$ to $1.04^3$. Since we have a large amount of training data, other data augmentation methods, like rotation, flipping, or color transformation are not utilized.

\subsubsection{Training Details}

We use an end-to-end method to train the proposed SiamVGG. We choose the SoftMargin loss as the loss function (Equation ~\ref{eq:loss}), where $y$ (defined in Equation~\ref{eq:y}) is the ground truth score map ($y[i] \in \{+1, -1\}$) and $x$ is the output score map. $n$ is the total number of elements in the score map.

\begin{align}\label{eq:loss}
    loss(x,y) = \sum_{i}\frac{log(1 + exp(-y[i] \times x[i]))}{n}
\end{align}

During training, we initialize the first ten convolutional layers of our design using VGG-16 pre-trained model on ILSVRC classification dataset, and use the method proposed in~\cite{he2015delving} to initialize the output layer. Stochastic gradient descent (SGD) is applied with learning rates from $10^{-4}$ to $10^{-7}$ during training. The whole training process contains over 200 epochs and each epoch consists of 6,000 sampled pairs. The gradients for each iteration are estimated using mini-batches of size 8. We use an IBM S822LC machine with 2$\times$Power8 CPUs, 512GB RAM, and 4$\times$Nvidia P100 GPUs to handle the training of SiamVGG.

\section{Experiments}
\label{sec:Experiments}

In this section, we first start an ablation study of our proposed SiamVGG on OTB100 dataset and then we demonstrate our approach in six different public datasets with the state-of-the-art performance. The implementation of our model is based on the PyTorch framework while the experiments (inferences) are running on a PC with an Intel i7-7700K, 16GB RAM, and Nvidia GTX 1080Ti GPU. 
We upsample the score map by using bicubic interpolation from $17 \times 17$ to $273 \times 273$ (16$\times$ upsampling with an element as the center of the score maps). To handle the scale transformation, we search for the object over three scales as $1.040^{-1,0,1}$ with the penalty rate of 0.97 for the scale variation and update the scale with a learning rate of 0.59.

\subsection{The OTB Benchmarks}

\begin{figure}
\centering
\includegraphics[width=.49\textwidth]{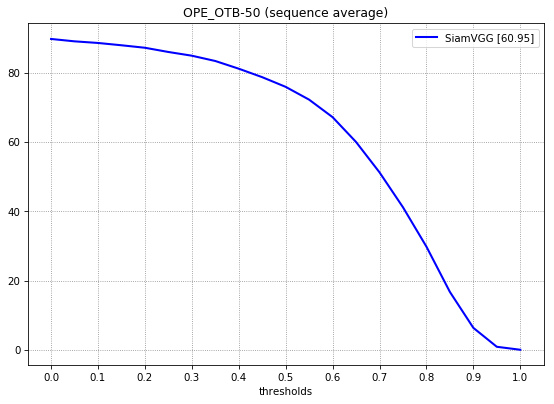}\hfill
\includegraphics[width=.49\textwidth]{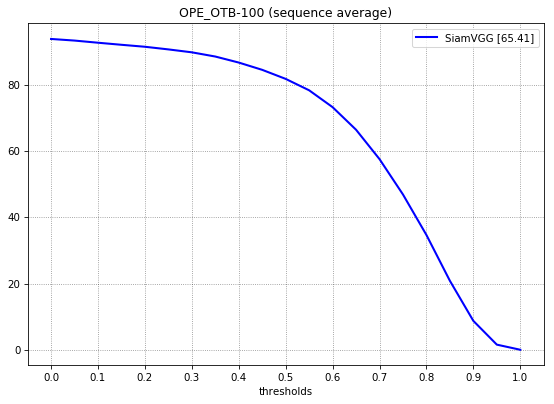}

\caption{Success plots for OPE (one pass evaluation) using OTB-50, OTB-100 benchmarks with the AUC value presented at the top-right corner in percentage.}

\label{fig:otb}
\end{figure}

We evaluate the proposed design in OTB benchmarks (OTB-2013, OTB-50, OTB-100), which are the most widely used public tracking benchmarks. These benchmarks consider the average per-frame success rate at different thresholds, which means a detection is considered to be successful in a given frame if the intersection-over-union (IoU) between its prediction result and the ground-truth is above a certain threshold. Trackers are then compared regarding area under the curve (AUC) of success rates for one pass evaluation (OPE). The results are shown in Figure~\ref{fig:otb} for OTB benchmarks along with different thresholds. Also, we compare our tracker to other state-of-the-art Siamese Network based trackers in Table~\ref{table:otb}. The results show that our proposed SiamVGG have achieved very competitive performance among all these trackers.

\setlength{\tabcolsep}{20pt}
\begin{table}
\begin{center}
\caption{AUC value for recently published real-time trackers using Siamese Networks. Datas highlighted in \textcolor{red}{\textit{red}}, \textcolor{blue}{\textit{blue}}, and \textcolor{green}{\textit{green}} color stand for the first, second, and third place of each benchmarks, respectively.}
\label{table:otb}
\begin{tabular}{cccc}
\hline
\textbf{tracker}&\textbf{OTB-2013} & \textbf{OTB-50} & \textbf{OTB-100} \\\hline

\textbf{SiamFC-3s~\cite{bertinetto2016fully}} & 0.607 & 0.516 & 0.582\\
\textbf{CFNet~\cite{Valmadre_2017_CVPR}} & 0.611 & \textcolor{green}{0.530} & 0.568\\
\textbf{RASNet~\cite{wang2018learning}} & \textcolor{blue}{0.670} & - & \textcolor{green}{0.642}  \\
\textbf{SA-Siam~\cite{he2018twofold}} & \textcolor{red}{0.677} & \textcolor{red}{0.610} & \textcolor{red}{0.657}  \\
\textbf{DSiam~\cite{guo2017learning}} & 0.656 & - & -  \\
\textbf{SiamRPN~\cite{li2018high}} & - & - & 0.637  \\\hline

\textbf{SiamVGG} & \textcolor{green}{0.665} & \textcolor{red}{0.610} & \textcolor{blue}{0.654} \\\hline

\end{tabular}
\end{center}
\end{table}
\setlength{\tabcolsep}{1.4pt}

\subsection{Ablation study on OTB100}

In this section, we explore several factors in the proposed design and eventually determine the final configuration of SiamVGG.

\subsubsection{Batch Normalization}

The first consideration relates to the batch normalization layer. Since the VGG-16 has two setups as one with batch normalization layer and the other without, we need to determine which is more suitable in tracking tasks. Although results on ILSVRC classification dataset show that the VGG-16 model with batch normalization can deliver relatively higher accuracy, we notice the other one (without batch normalization) achieves much higher performance in tracking tasks with 0.654 AUC of success plots for OPE than the model with batch normalization (with 0.589 AUC).

\subsubsection{Youtube-BB Dataset}

We notice that the ILSVRC only contains less than 5,000 video sequences, which are inadequate and easily cause overfitting during training. Thus, we introduce the Youtube-BB dataset including more than 100,000 videos annotated once in every 30 frames for more training material. By combining these two datasets in a particular ratio, the performance of our design has improved from 0.637 to 0.654 (AUC of success plots for OPE) in OTB-100.

\subsection{The VOT Benchmarks}

We use the latest version of the Visual Object Tracking toolkit (6.0.3) to start reset-based experiments. The toolkit re-initializes the tracker in five frames after failure is detected (zero overlaps with the ground truth). The performance is measured in terms of expected average overlap (EAO) to reflect both robustness and accuracy quantitatively. For the comparison in VOT2017 participants, we also include the real-time tracking performance as the additional metrics. Comparison results to other state-of-the-art methods on VOT2015, VOT2016, and VOT2017 are reported in the following subsections.
\begin{figure}
\centering
\includegraphics[width=3cm, height=1.6875cm]{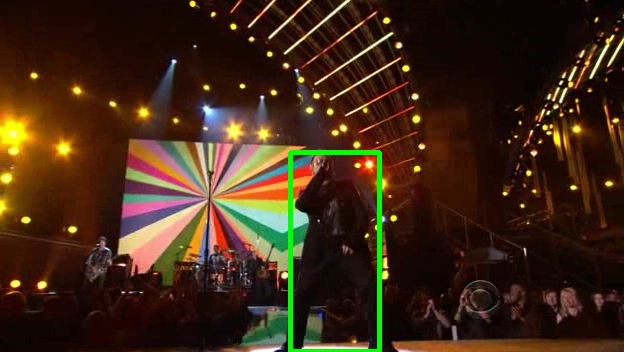}\hfill
\includegraphics[width=3cm, height=1.6875cm]{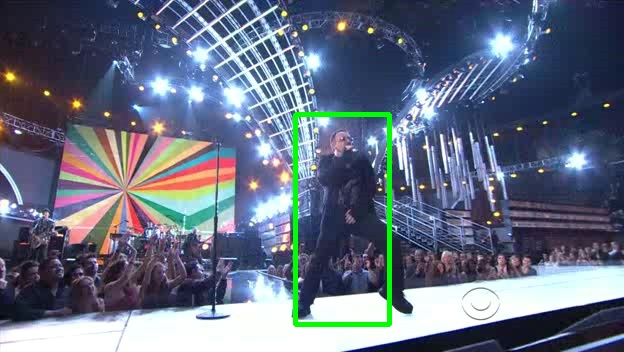}\hfill
\includegraphics[width=3cm, height=1.6875cm]{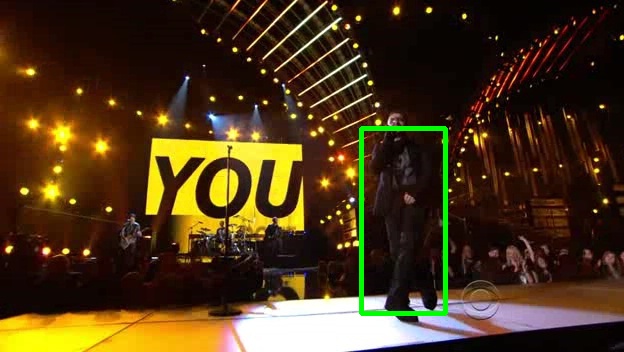}\hfill
\includegraphics[width=3cm, height=1.6875cm]{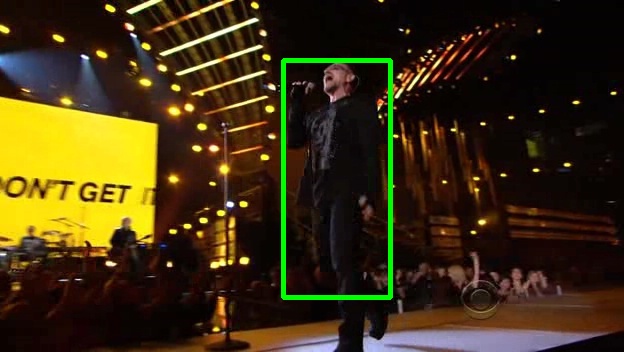}\hfill\\
\includegraphics[width=3cm, height=1.6875cm]{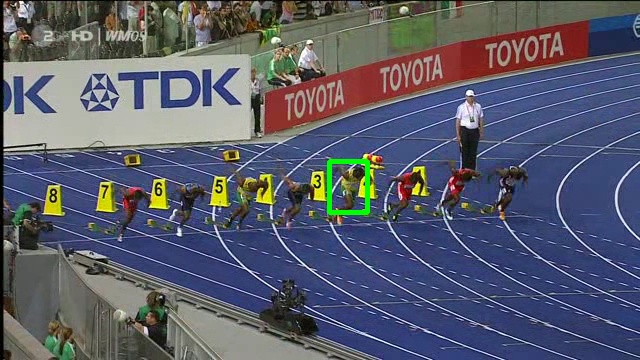}\hfill
\includegraphics[width=3cm, height=1.6875cm]{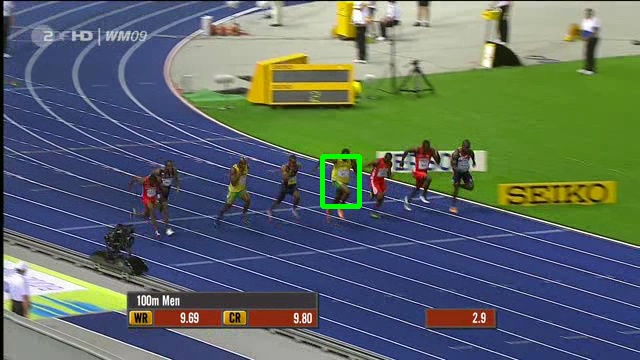}\hfill
\includegraphics[width=3cm, height=1.6875cm]{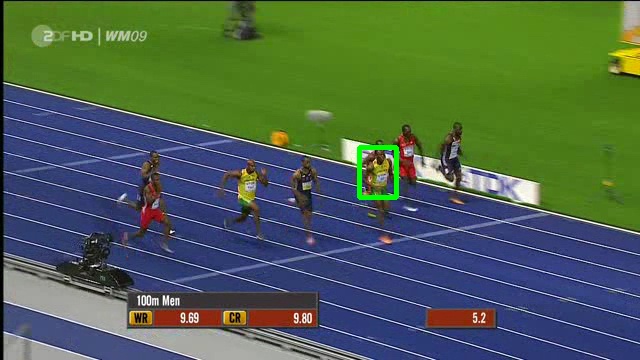}\hfill
\includegraphics[width=3cm, height=1.6875cm]{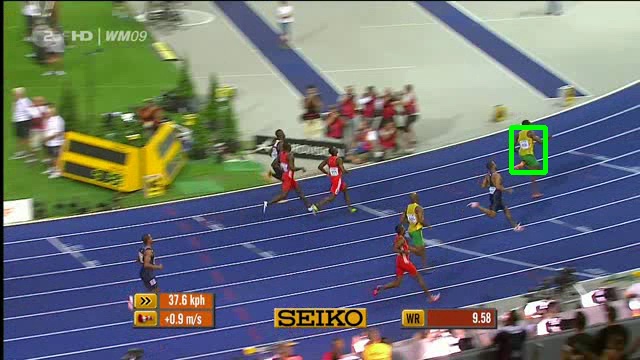}\hfill\\
\includegraphics[width=3cm, height=1.6875cm]{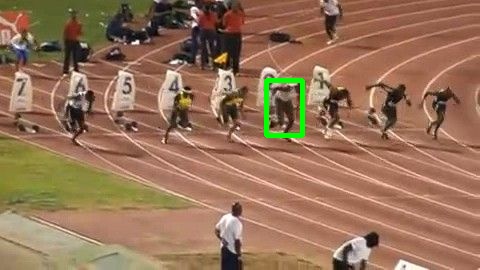}\hfill
\includegraphics[width=3cm, height=1.6875cm]{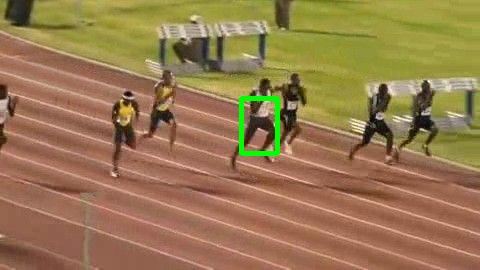}\hfill
\includegraphics[width=3cm, height=1.6875cm]{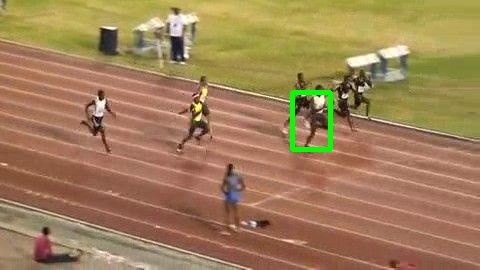}\hfill
\includegraphics[width=3cm, height=1.6875cm]{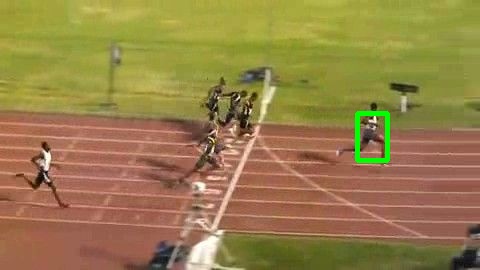}\hfill\\
\includegraphics[width=3cm, height=1.6875cm]{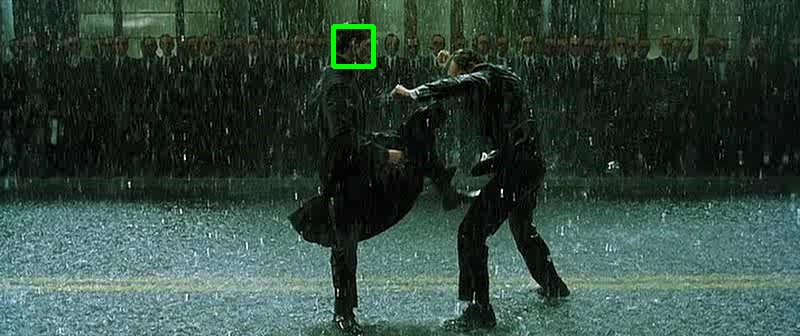}\hfill
\includegraphics[width=3cm, height=1.6875cm]{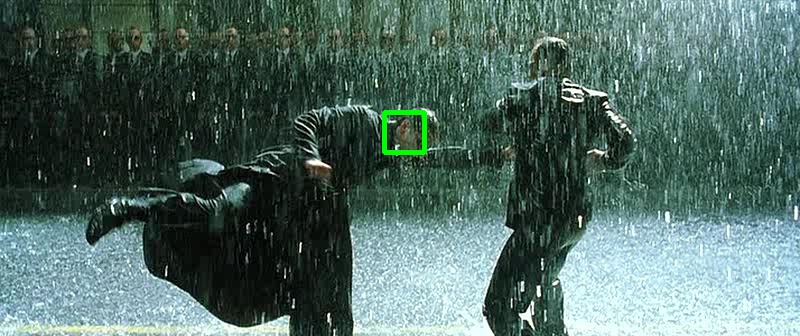}\hfill
\includegraphics[width=3cm, height=1.6875cm]{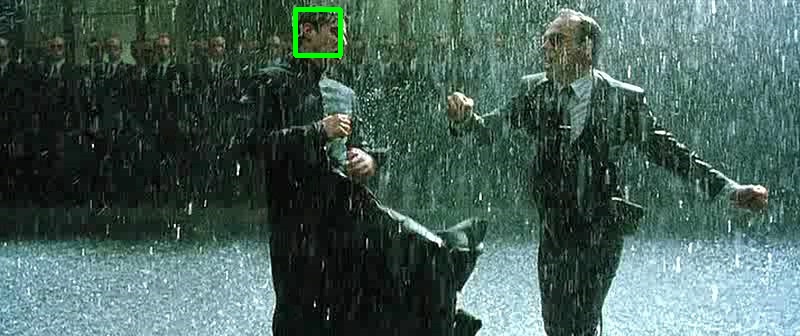}\hfill
\includegraphics[width=3cm, height=1.6875cm]{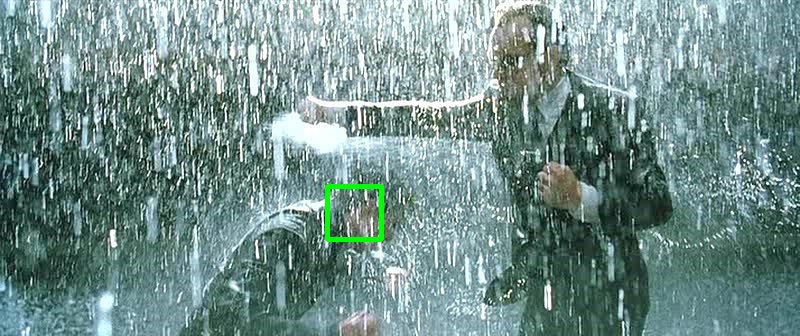}\hfill\\
\includegraphics[width=3cm, height=1.6875cm]{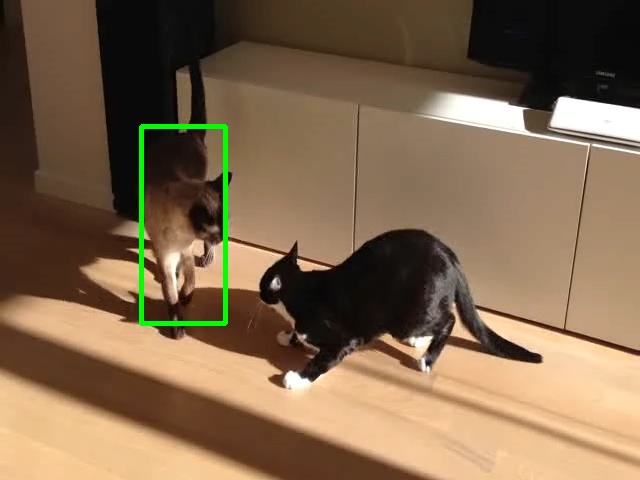}\hfill
\includegraphics[width=3cm, height=1.6875cm]{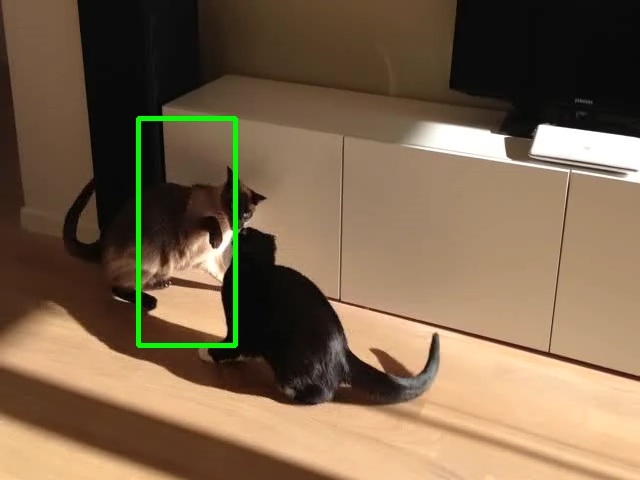}\hfill
\includegraphics[width=3cm, height=1.6875cm]{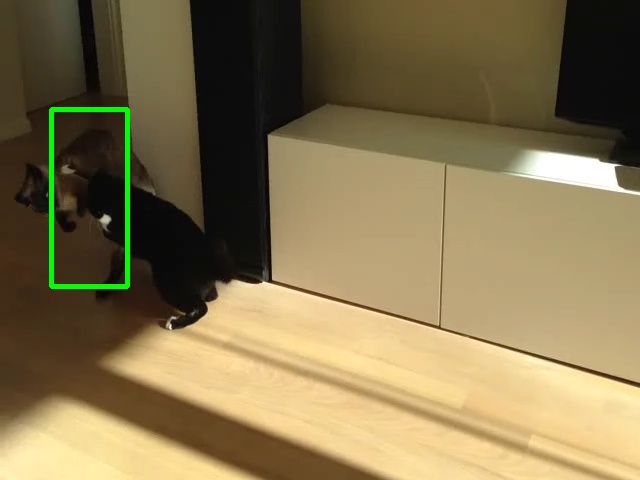}\hfill
\includegraphics[width=3cm, height=1.6875cm]{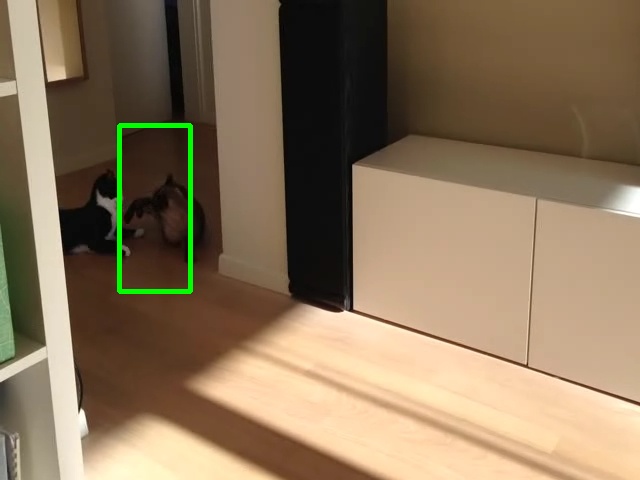}\hfill

\vspace{-8pt}
    
\caption{Snapshots of the results generated by the proposed SiamVGG on VOT2017. From top row to the bottom row are~\textit{singer3},~\textit{bolt1},~\textit{bolt2},~\textit{matrix},~\textit{fernando}, respectively.}

\label{fig:xzgt}
\end{figure}

\subsubsection{Results on VOT2015}

The VOT2015 dataset consists of 60 sequences. The performance is evaluated in terms of accuracy (average overlap while tracking successfully) and robustness (failure times). The overall performance is measured using an expected average overlap (EAO) which takes account of both accuracy and robustness quantitatively. 

We compare our tracker with top 10 trackers recorded in the VOT2015 presentation (remove MDNet from the record because it's trained with data generated from OTB's sequences). Also, we compare our tracker with other previous methods based on the Siamese Networks (if their results are reported). 
The result in Table~\ref{table:vot2015} shows that SiamVGG is able to rank \textit{1st} in EAO with 17\% enhancement compared to the DeepSRDCF which cannot run at real-time speed. Although our tracker is slower than SiamRPN, our method's EAO is relatively higher.

\setlength{\tabcolsep}{20pt}
\begin{table}
\begin{center}
\caption{We compare our tracker to top 10 trackers recorded in the VOT2015 challenge (MDNet~\cite{nam2016learning} is removed) and several Siamese Network based methods regarding overlap, failures, and expected average overlap (EAO) using the latest VOT toolkit (6.0.3).}
\label{table:vot2015}
\begin{tabular}{cccc}
\hline

\textbf{Tracker}&\textbf{Overlap} & \textbf{Failures} & \textbf{EAO} \\\hline

\textbf{DeepSRDCF~\cite{danelljan2015convolutional}} & 0.556 & \textcolor{green}{16.953} & 0.318\\
\textbf{EBT~\cite{zhu2015tracking}} & 0.459 & \textcolor{blue}{15.370} & 0.313 \\
\textbf{LDP} & 0.484 & 23.897 & 0.278  \\
\textbf{NSAMF} & 0.525 & 25.616 & 0.254 \\
\textbf{RAJSSC} & \textcolor{green}{0.563} & 29.761 & 0.242\\
\textbf{S3Tracker} & 0.505 & 27.856 & 0.240\\
\textbf{SC-EBT} &0.542 & 31.816 & 0.255  \\
\textbf{sPST~\cite{hua2015online}} & 0.548 & 26.253 & 0.277  \\
\textbf{SRDCF~\cite{danelljan2015learning}} & 0.553 & 21.264 & 0.288  \\
\textbf{Struck~\cite{hare2016struck}} & 0.454 & 27.153 & 0.246  \\\hline

\textbf{RasNet~\cite{wang2018learning}} & - & - & \textcolor{green}{0.327}  \\

\textbf{SA-Siam~\cite{he2018twofold}} & - & - & 0.310  \\

\textbf{SiamFC-3s~\cite{bertinetto2016fully}} & 0.534 & - & 0.289  \\

\textbf{SiamFC-5s~\cite{bertinetto2016fully}} & 0.524 & - & 0.274  \\

\textbf{SiamRPN~\cite{li2018high}} &\textcolor{blue}{ 0.580} & - & \textcolor{blue}{0.358}  \\\hline

\textbf{SiamVGG} &\textcolor{red}{0.601} & \textcolor{red}{12.506} & \textcolor{red}{0.373}\\\hline

\end{tabular}
\end{center}
\end{table}
\setlength{\tabcolsep}{1.4pt}

\subsubsection{Results on VOT2016}

The VOT2016 dataset keeps the same sequences as the VOT2015 with re-annotated bounding boxes. We use the same performance evaluation methods as VOT2015 and compare our tracker to the top 10 trackers in VOT2016 and those using Siamese Networks. The result in Table~\ref{table:vot2016} shows that our tracker achieves the first place in both failures and EAO sections, and the second place regarding overlap section. Our tracker can also beat the C-COT with better real-time speed and deliver significant improvement compared to the advanced Siamese Network based methods.

\setlength{\tabcolsep}{20pt}
\begin{table}
\begin{center}
\caption{We compare our tracker to top 10 trackers recorded in the VOT2016 challenge and several Siamese Network based methods regarding overlap, failures, expected average overlap (EAO) using the latest VOT toolkit (6.0.3).}
\label{table:vot2016}
\begin{tabular}{cccc}
\hline

\textbf{Tracker}&\textbf{Overlap} & \textbf{Failures} & \textbf{EAO} \\\hline

\textbf{C-COT~\cite{danelljan2016beyond}} & 0.533 & 16.582 &\textcolor{green}{ 0.331}  \\
\textbf{DDC} & 0.534 & 20.981 & 0.293 \\
\textbf{DNT} & 0.509 & 19.544 & 0.278  \\
\textbf{EBT~\cite{zhu2015tracking}} & 0.453 & \textcolor{green}{15.194} & 0.291 \\
\textbf{MLDF} & 0.487 & \textcolor{blue}{15.044} & 0.311\\
\textbf{SRBT} & 0.484 & 21.325 & 0.290\\
\textbf{SSAT} & \textcolor{red}{0.570} & 19.272 & 0.321  \\
\textbf{Staple~\cite{bertinetto2016staple}} & 0.543 & 23.895 & 0.295  \\
\textbf{STAPLE+} & 0.552 & 24.316 & 0.286  \\
\textbf{TCNN~\cite{nam2016modeling}} & 0.547 & 17.939 & 0.325  \\\hline

\textbf{SA-Siam~\cite{he2018twofold}} & - & - & 0.291  \\

\textbf{SiamFC-A} & - & - & 0.235  \\
\textbf{SiamFC-R} & - & - & 0.277  \\

\textbf{SiamRPN~\cite{li2018high}} & \textcolor{green}{0.560} & - & \textcolor{blue}{0.344}  \\\hline

\textbf{SiamVGG} & \textcolor{blue}{0.564} & \textcolor{red}{14.328} & \textcolor{red}{0.351} \\\hline

\end{tabular}
\end{center}
\end{table}
\setlength{\tabcolsep}{1.4pt}

\subsubsection{Results on VOT2017}

In VOT2017, the easiest 10 sequences in dataset are replaced by updated sequences. In addition, specific real-time performance is required, where trackers need to run on at least 25 FPS. If the tracker fails to generate each result in 40 ms (25 FPS), the VOT toolkit will keep the bounding box of the last frame as the result of the current frame. As shown in Table~\ref{table:vot2017}, our tracker achieves the first place in overlap and the third place in EAO section, respectively. Note that most of the top 10 trackers listed in Table~\ref{table:vot2017} cannot run at real-time speed while maintaining good accuracy.

\begin{figure}[h]
\centering
\includegraphics[height=6.5cm]{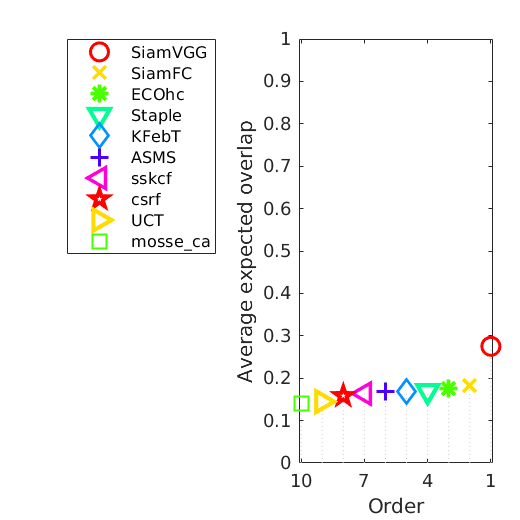}
\caption{The real-time expected average overlap analysis for SiamVGG and other real-time trackers which recorded in the VOT2017 real-time challenge.}

\label{fig:vot2017realtime}
\end{figure}

To evaluate the real-time performance, we compare our method to other state-of-the-art real-time trackers reported in VOT2017 and the Siamese Network based trackers and show the results in Table~\ref{table:vot2017realtime}. It indicates that our tracker can achieve significant improvement on real-time with 13\% higher EAO compared to SiamRPN. 
In VOT2017, a toolkit developed in Matlab is used to evaluate the FPS performance. In our case, we collect the FPS results by converting the toolkit API from Matlab to Python and present them in Table~\ref{table:vot2017realtime}. Note that there is a certain number of FPS drop due to the API conversion. 
In our case, the proposed SiamVGG can reach 50 FPS using a GTX 1080Ti for tracking tasks without evaluating by the VOT toolkit. By comparing to the baseline model, SiamFC, our tracker can deliver 51\% better EAO without significant speed loss.
We also visualize the real-time EAO of all selected trackers in Figure~\ref{fig:vot2017realtime}.

\setlength{\tabcolsep}{20pt}
\begin{table}
\begin{center}
\caption{We compare our tracker to top 10 trackers recorded in the VOT2017 challenge and several Siamese Network based methods regarding overlap, failures, expected average overlap (EAO) using the latest VOT toolkit (6.0.3).}
\label{table:vot2017}
\begin{tabular}{cccc}
\hline

\textbf{Tracker}&\textbf{Overlap} & \textbf{Failures} & \textbf{EAO} \\\hline

\textbf{C-COT~\cite{danelljan2016beyond}} & 0.485 & 20.414 & 0.267  \\
\textbf{CFCF~\cite{danelljan2015convolutional}} & \textcolor{green}{0.505} & 19.649 & \textcolor{green}{0.286} \\
\textbf{CFWCR} & 0.483 & \textcolor{blue}{17.134} & \textcolor{blue}{0.303}  \\
\textbf{CSRDCF~\cite{lukezic2017discriminative}} & - & - & 0.256\\
\textbf{ECO~\cite{danelljan2017eco}} & 0.476 & 17.663 & 0.280 \\
\textbf{Gnet} & 0.500 & \textcolor{green}{17.367} & 0.274\\
\textbf{LSART~\cite{sun2018learning}} & 0.490 & \textcolor{red}{12.793} & \textcolor{red}{0.323}  \\
\textbf{MCCT~\cite{wang2018multi}} & \textcolor{blue}{0.523} & 19.453 & 0.270  \\
\textbf{MCPF~\cite{zhang2017multi}} & 0.503 & 25.960 & 0.248  \\
\textbf{SiamDCF} & 0.496 & 29.406 & 0.249  \\\hline

\textbf{RasNet~\cite{wang2018learning}} & - & - & 0.281  \\

\textbf{SA-Siam~\cite{he2018twofold}} & - & - & 0.236  \\
\textbf{SiamFC~\cite{bertinetto2016fully}} & - & - & 0.188  \\\hline

\textbf{SiamVGG} & \textcolor{red}{0.525} & 20.453 & \textcolor{green}{0.286} \\\hline
\end{tabular}
\end{center}
\end{table}
\setlength{\tabcolsep}{1.4pt}

\setlength{\tabcolsep}{30pt}
\begin{table}[h!]
\begin{center}
\caption{We compare our tracker to top 10 trackers recorded in the VOT2017 real-time challenge and several Siamese Network based methods regarding real-time expected average overlap (EAO) and frames per second (FPS) using the latest VOT toolkit (6.0.3).}
\label{table:vot2017realtime}
\begin{tabular}{cccc}
\hline

\textbf{Tracker}&\textbf{EAO} & \textbf{FPS}  \\\hline
\textbf{ASMS} & 0.168 & \textcolor{blue}{133.28}  \\
\textbf{CSRDCF++} & 0.212 & -  \\
\textbf{csrf} & 0.158 &16.41 \\

\textbf{ECOhc~\cite{danelljan2017eco}} & 0.177 & 21.44  \\
\textbf{KFebT~\cite{SennaDrumBast:2017:ReEnTr}} & 0.169 & \textcolor{red}{138.95} \\
\textbf{mosse\_ca} & 0.139 & \textcolor{green}{62.45}  \\
\textbf{SiamFC~\cite{bertinetto2016fully}} & 0.182 & 35.24   \\

\textbf{sskcf} & 0.164 & 45.40 \\

\textbf{Staple~\cite{bertinetto2016staple}} & 0.170 & 54.26   \\

\textbf{UCT~\cite{zhu2017uct}} & 0.145 & 14.73\\\hline

\textbf{RasNet~\cite{wang2018learning}} & 0.223 & - \\

\textbf{SA-Siam~\cite{he2018twofold}} & \textcolor{green}{0.236}& - \\
\textbf{SiamRPN~\cite{li2018high}} & \textcolor{blue}{0.243} & -  \\\hline

\textbf{SiamVGG} &\textcolor{red}{ 0.275} & 33.15  \\\hline
\end{tabular}
\end{center}
\end{table}
\setlength{\tabcolsep}{1.4pt}

\section{Conclusion}
\label{sec:conclusion}
In this work, we proposed a SiamVGG tracker which is an end-to-end DNN-based model featuring offline parameter update from large-scale image pairs (ILSVRC and Youtube-BB datasets). By modifying the network from the baseline SiamFC method, our proposed SiamVGG can deliver significant improvements of tracking performance with the state-of-the-art real-time EAO. SiamVGG is very easy to reduplicate and can be deployed onto IoT devices (with limited computation and memory resources) because of its compact network structure. Experiments showed that our method can outperform most the existing trackers evaluated on VOT2015/2016/2017 datasets and OTB-2013/50/100 datasets and we can maintain very high FPS for real-time tracking.

\section{Acknowledgement}
We thank the IBM-Illinois Center for Cognitive Computing Systems Research (C3SR) – a research collaboration as part of IBM AI Horizons Network for supporting this research.


\bibliographystyle{splncs}

\begin{thebibliography}{10}

\bibitem{danelljan2017eco}
Danelljan, M., Bhat, G., Khan, F.S., Felsberg, M.:
\newblock Eco: Efficient convolution operators for tracking.
\newblock In: Proceedings of the 2017 IEEE Conference on Computer Vision and
  Pattern Recognition (CVPR), Honolulu, HI, USA. (2017)  21--26

\bibitem{danelljan2015convolutional}
Danelljan, M., Hager, G., Shahbaz~Khan, F., Felsberg, M.:
\newblock Convolutional features for correlation filter based visual tracking.
\newblock In: Proceedings of the IEEE International Conference on Computer
  Vision Workshops. (2015)  58--66

\bibitem{nvdawhitepaper1}
Nvidia:
\newblock Gpu-based deep learning inference.
\newblock In: Nvidia White paper. (2015)

\bibitem{nvdawhitepaper2}
Franklin, D.:
\newblock Nvidia {Jetson} {TX2} delivers twice the intelligence to the edge.
\newblock In: Nvidia White paper. (2017)

\bibitem{zhang2017high}
Zhang, X., Liu, X., Ramachandran, A., Zhuge, C., Tang, S., Ouyang, P., Cheng,
  Z., Rupnow, K., Chen, D.:
\newblock High-performance video content recognition with long-term recurrent
  convolutional network for fpga.
\newblock In: Field Programmable Logic and Applications (FPL), 2017 27th
  International Conference on, IEEE (2017)  1--4

\bibitem{zhangdnnbuilder}
Zhang, X., Wang, J., Zhu, C., Lin, Y., Xiong, J., Hwu, W.m., Chen, D.:
\newblock Dnnbuilder: an automated tool for building high-performance dnn
  hardware accelerators for fpgas.
\newblock In: Proceedings of the 37th International Conference on
  Computer-Aided Design, IEEE (2018)

\bibitem{isscc_2016_chen_eyeriss}
Chen, Y. H., Krishna, T., Emer, J. S., Sze, V:
\newblock Eyeriss: An Energy-Efficient Reconfigurable Accelerator for Deep Convolutional Neural Networks.
\newblock In: IEEE journal of solid-state circuits, (2016) 127--138

\bibitem{han2016eie}
Han, S., Liu, X., Mao, H., et~al.:
\newblock EIE: Efficient inference engine on compressed deep neural network.
\newblock In: ACM SIGARCH Computer Architecture News, (2016) 243-254

\bibitem{jouppi2017datacenter}
Jouppi, N.P., Young, C., Patil, N.,  et~al.:
\newblock In-datacenter performance analysis of a tensor processing unit.
\newblock In: Computer Architecture (ISCA), 2017 ACM/IEEE 44th Annual
  International Symposium on, IEEE (2017)  1--12

\bibitem{bolme2010visual}
Bolme, D.S., Beveridge, J.R., Draper, B.A., Lui, Y.M.:
\newblock Visual object tracking using adaptive correlation filters.
\newblock In: Computer Vision and Pattern Recognition (CVPR), 2010 IEEE
  Conference on, IEEE (2010)  2544--2550

\bibitem{krizhevsky2012imagenet}
Krizhevsky, A., Sutskever, I., Hinton, G.E.:
\newblock Imagenet classification with deep convolutional neural networks.
\newblock In: Advances in neural information processing systems. (2012)
  1097--1105

\bibitem{redmon2017yolo9000}
Redmon, J., Farhadi, A.:
\newblock Yolo9000: better, faster, stronger.
\newblock arXiv preprint (2017)

\bibitem{li2018csrnet}
Li, Y., Zhang, X., Chen, D.:
\newblock Csrnet: Dilated convolutional neural networks for understanding the
  highly congested scenes.
\newblock In: Proceedings of the IEEE Conference on Computer Vision and Pattern
  Recognition. (2018)  1091--1100

\bibitem{bertinetto2016fully}
Bertinetto, L., Valmadre, J., Henriques, J.F., Vedaldi, A., Torr, P.H.:
\newblock Fully-convolutional siamese networks for object tracking.
\newblock In: European conference on computer vision, Springer (2016)  850--865

\bibitem{simonyan2014very}
Simonyan, K., Zisserman, A.:
\newblock Very deep convolutional networks for large-scale image recognition.
\newblock arXiv preprint arXiv:1409.1556 (2014)

\bibitem{real2017youtube}
Real, E., Shlens, J., Mazzocchi, S., Pan, X., Vanhoucke, V.:
\newblock Youtube-boundingboxes: A large high-precision human-annotated data
  set for object detection in video.
\newblock In: 2017 IEEE Conference on Computer Vision and Pattern Recognition
  (CVPR), IEEE (2017)  7464--7473

\bibitem{WuLimYang13}
Wu, Y., Lim, J., Yang, M.H.:
\newblock Online object tracking: A benchmark.
\newblock In: Proceedings of the IEEE conference on computer vision and pattern
  recognition. (2013)  2411--2418

\bibitem{Kristan2017a}
Kristan, M., Leonardis, A., Matas, J., Felsberg, M., Pflugfelder, R.,
  \v{C}ehovin Zajc, L., Vojir, T., H\"{a}ger, G., Luke\v{z}i\v{c}, A.,
  Eldesokey, A., Fernandez, G.:
\newblock The visual object tracking vot2017 challenge results (2017)

\bibitem{VOT_TPAMI}
Kristan, M., Matas, J., Leonardis, A., Voj{\'\i}{\v{r}}, T., Pflugfelder, R.,
  Fernandez, G., Nebehay, G., Porikli, F., {\v{C}}ehovin, L.:
\newblock A novel performance evaluation methodology for single-target
  trackers.
\newblock IEEE transactions on pattern analysis and machine intelligence
  \textbf{38}(11) (2016)  2137--2155

\bibitem{danelljan2016beyond}
Danelljan, M., Robinson, A., Khan, F.S., Felsberg, M.:
\newblock Beyond correlation filters: Learning continuous convolution operators
  for visual tracking.
\newblock In: European Conference on Computer Vision, Springer (2016)  472--488

\bibitem{huang2017learning}
Huang, C., Lucey, S., Ramanan, D.:
\newblock Learning policies for adaptive tracking with deep feature cascades.
\newblock arXiv preprint arXiv:1708.02973 (2017)

\bibitem{guo2017learning}
Guo, Q., Feng, W., Zhou, C., Huang, R., Wan, L., Wang, S.:
\newblock Learning dynamic siamese network for visual object tracking.
\newblock In: Proc. IEEE Int. Conf. Comput. Vis. (2017)  1--9

\bibitem{he2018twofold}
He, A., Luo, C., Tian, X., Zeng, W.:
\newblock A twofold siamese network for real-time object tracking.
\newblock In: Proceedings of the IEEE Conference on Computer Vision and Pattern
  Recognition. (2018)  4834--4843

\bibitem{wang2018learning}
Wang, Q., Teng, Z., Xing, J., Gao, J., Hu, W., Maybank, S.:
\newblock Learning attentions: residual attentional siamese network for high
  performance online visual tracking.
\newblock In: Proceedings of the IEEE Conference on Computer Vision and Pattern
  Recognition. (2018)  4854--4863

\bibitem{li2018high}
Li, B., Yan, J., Wu, W., Zhu, Z., Hu, X.:
\newblock High performance visual tracking with siamese region proposal
  network.
\newblock In: Proceedings of the IEEE Conference on Computer Vision and Pattern
  Recognition. (2018)  8971--8980

\bibitem{long2015fully}
Long, J., Shelhamer, E., Darrell, T.:
\newblock Fully convolutional networks for semantic segmentation.
\newblock In: Proceedings of the IEEE conference on computer vision and pattern
  recognition. (2015)  3431--3440

\bibitem{he2015delving}
He, K., Zhang, X., Ren, S., Sun, J.:
\newblock Delving deep into rectifiers: Surpassing human-level performance on
  imagenet classification.
\newblock In: Proceedings of the IEEE international conference on computer
  vision. (2015)  1026--1034

\bibitem{Valmadre_2017_CVPR}
Valmadre, J., Bertinetto, L., Henriques, J., Vedaldi, A., Torr, P.H.S.:
\newblock End-to-end representation learning for correlation filter based
  tracking.
\newblock In: The IEEE Conference on Computer Vision and Pattern Recognition
  (CVPR). (July 2017)

\bibitem{nam2016learning}
Nam, H., Han, B.:
\newblock Learning multi-domain convolutional neural networks for visual
  tracking.
\newblock In: Computer Vision and Pattern Recognition (CVPR), 2016 IEEE
  Conference on, IEEE (2016)  4293--4302

\bibitem{zhu2015tracking}
Zhu, G., Porikli, F., Li, H.:
\newblock Tracking randomly moving objects on edge box proposals.
\newblock arXiv preprint arXiv:1507.08085 (2015)

\bibitem{hua2015online}
Hua, Y., Alahari, K., Schmid, C.:
\newblock Online object tracking with proposal selection.
\newblock In: Proceedings of the IEEE international conference on computer
  vision. (2015)  3092--3100

\bibitem{danelljan2015learning}
Danelljan, M., Hager, G., Shahbaz~Khan, F., Felsberg, M.:
\newblock Learning spatially regularized correlation filters for visual
  tracking.
\newblock In: Proceedings of the IEEE International Conference on Computer
  Vision. (2015)  4310--4318

\bibitem{hare2016struck}
Hare, S., Golodetz, S., Saffari, A., Vineet, V., Cheng, M.M., Hicks, S.L.,
  Torr, P.H.:
\newblock Struck: Structured output tracking with kernels.
\newblock IEEE transactions on pattern analysis and machine intelligence
  \textbf{38}(10) (2016)  2096--2109

\bibitem{bertinetto2016staple}
Bertinetto, L., Valmadre, J., Golodetz, S., Miksik, O., Torr, P.H.:
\newblock Staple: Complementary learners for real-time tracking.
\newblock In: Proceedings of the IEEE conference on computer vision and pattern
  recognition. (2016)  1401--1409

\bibitem{nam2016modeling}
Nam, H., Baek, M., Han, B.:
\newblock Modeling and propagating cnns in a tree structure for visual
  tracking.
\newblock arXiv preprint arXiv:1608.07242 (2016)

\bibitem{lukezic2017discriminative}
Lukezic, A., Vojir, T., Zajc, L.C., Matas, J., Kristan, M.:
\newblock Discriminative correlation filter with channel and spatial
  reliability.
\newblock In: CVPR. Volume~6. (2017) ~8

\bibitem{sun2018learning}
Sun, C., Lu, H., Yang, M.H.:
\newblock Learning spatial-aware regressions for visual tracking.
\newblock In: IEEE Conf. on Computer Vision and Pattern Recognition (CVPR).
  (2018)  8962--8970

\bibitem{wang2018multi}
Wang, N., Zhou, W., Tian, Q., Hong, R., Wang, M., Li, H.:
\newblock Multi-cue correlation filters for robust visual tracking.
\newblock In: Proceedings of the IEEE Conference on Computer Vision and Pattern
  Recognition. (2018)  4844--4853

\bibitem{zhang2017multi}
Zhang, T., Xu, C., Yang, M.H.:
\newblock Multi-task correlation particle filter for robust object tracking.
\newblock In: Proceedings of the IEEE Conference on Computer Vision and Pattern
  Recognition. Volume~1. (2017) ~3

\bibitem{SennaDrumBast:2017:ReEnTr}
Senna, P., Drummond, I.N., Bastos, G.S.:
\newblock Real-time ensemble-based tracker with kalman filter.
\newblock In: Electronic Proceedings of the 30th Conference on Graphics,
  Patterns and Images (SIBGRAPI'17), Conference on Graphics, Patterns and
  Images, 30.(SIBGRAPI) (2017)

\bibitem{zhu2017uct}
Zhu, Z., Huang, G., Zou, W., Du, D., Huang, C.:
\newblock Uct: learning unified convolutional networks for real-time visual
  tracking.
\newblock In: Proc. of the IEEE Int. Conf. on Computer Vision Workshops. (2017)
   1973--1982

\bibitem{xu2020autodnnchip}Xu, P., Zhang, X., Hao, C., Zhao, Y., Zhang, Y., Wang, Y., Li, C., Guan, Z., Chen, D. \& Lin, Y. AutoDNNchip: An automated dnn chip predictor and builder for both FPGAs and ASICs. {\em Proceedings Of The 2020 ACM/SIGDA International Symposium On Field-Programmable Gate Arrays}. pp. 40-50 (2020)

\bibitem{li2020edd}Li, Y., Hao, C., Zhang, X., Liu, X., Chen, Y., Xiong, J., Hwu, W. \& Chen, D. Edd: Efficient differentiable dnn architecture and implementation co-search for embedded ai solutions. {\em 2020 57th ACM/IEEE Design Automation Conference (DAC)}. pp. 1-6 (2020)

\bibitem{zhang2018dnnbuilder}Zhang, X., Wang, J., Zhu, C., Lin, Y., Xiong, J., Hwu, W. \& Chen, D. DNNBuilder: An automated tool for building high-performance DNN hardware accelerators for FPGAs. {\em 2018 IEEE/ACM International Conference On Computer-Aided Design (ICCAD)}. pp. 1-8 (2018)

\bibitem{chen2016platform}Chen, D., Cong, J., Gurumani, S., Hwu, W., Rupnow, K. \& Zhang, Z. Platform choices and design demands for IoT platforms: cost, power, and performance tradeoffs. {\em IET Cyber-Physical Systems: Theory \& Applications}. \textbf{1}, 70-77 (2016)

\bibitem{liu2011real}Liu, S., Papakonstantinou, A., Wang, H. \& Chen, D. Real-time object tracking system on FPGAs. {\em 2011 Symposium On Application Accelerators In High-Performance Computing}. pp. 1-7 (2011)

\bibitem{zhao2012real}Zhao, S., Ahmed, S., Liang, Y., Rupnow, K., Chen, D. \& Jones, D. A real-time 3D sound localization system with miniature microphone array for virtual reality. {\em 2012 7th IEEE Conference On Industrial Electronics And Applications (ICIEA)}. pp. 1853-1857 (2012)

\bibitem{zhuge2018face}Zhuge, C., Liu, X., Zhang, X., Gummadi, S., Xiong, J. \& Chen, D. Face recognition with hybrid efficient convolution algorithms on FPGAs. {\em Proceedings Of The 2018 On Great Lakes Symposium On VLSI}. pp. 123-128 (2018)

\bibitem{zhang2020skynet}Zhang, X., Lu, H., Hao, C., Li, J., Cheng, B., Li, Y., Rupnow, K., Xiong, J., Huang, T., Shi, H. \& Others SkyNet: a hardware-efficient method for object detection and tracking on embedded systems. {\em Proceedings Of Machine Learning And Systems}. \textbf{2} pp. 216-229 (2020)

\bibitem{he2009novel}He, C., Papakonstantinou, A. \& Chen, D. A novel SoC architecture on FPGA for ultra fast face detection. {\em 2009 IEEE International Conference On Computer Design}. pp. 412-418 (2009)

\end{thebibliography}

\end{document}